\def\year{2021}\relax
\documentclass[letterpaper]{article} 
\usepackage{aaai21}  
\usepackage{times}  
\usepackage{helvet} 
\usepackage{courier}  
\usepackage[hyphens]{url}  
\usepackage{graphicx} 
\usepackage{natbib}  
\usepackage{caption} 
\usepackage{graphicx}
\usepackage{amsmath}
\usepackage{booktabs}
\usepackage{array}
\usepackage{amsfonts}
\usepackage{bbm}
\usepackage{enumerate}
\newcommand{\bertbase}{BERT$_{\textrm{BASE}}$}
\newcommand{\bertlarge}{BERT$_{\textrm{LARGE}}$}
\newcommand{\robertalarge}{RoBERTa$_{\textrm{LARGE}}$}

\title{REM-Net: Recursive Erasure Memory Network \\for Commonsense Evidence Refinement}
\author{
    Yinya Huang\textsuperscript{\rm 1}, Meng Fang\textsuperscript{\rm 2}, 
    Xunlin Zhan\textsuperscript{\rm 1}, Qingxing Cao\textsuperscript{\rm 1},
    Xiaodan Liang\textsuperscript{\rm 1}\thanks{Corresponding author: Xiaodan Liang.},
    Liang Lin\textsuperscript{\rm 1} \\
}
\affiliations{
    \textsuperscript{\rm 1} Shenzhen Campus of Sun Yat-sen University \\ 
    \textsuperscript{\rm 2} Tencent AI Lab / Robotics X \\
    yinya.huang@hotmail.com, mfang@tencent.com, \\
    \{zhanxlin, caoqx\}@mail2.sysu.edu.cn, xdliang328@gmail.com, linliang@ieee.org
}

\begin{document}

\maketitle

\begin{abstract}
When answering a question, people often draw upon their rich world knowledge in addition to the particular context. While recent works retrieve supporting facts/evidence from commonsense knowledge bases to supply additional information to each question, there is still ample opportunity to advance it on the quality of the evidence. It is crucial since the quality of the evidence is the key to answering commonsense questions, and even determines the upper bound on the QA systems' performance. In this paper, we propose a recursive erasure memory network (REM-Net) to cope with the quality improvement of evidence. To address this, REM-Net is equipped with a module to refine the evidence by recursively erasing the low-quality evidence that does not explain the question answering. Besides, instead of retrieving evidence from existing knowledge bases, REM-Net leverages a pre-trained generative model to generate candidate evidence customized for the question. We conduct experiments on two commonsense question answering datasets, WIQA and CosmosQA. The results demonstrate the performance of REM-Net and show that the refined evidence is explainable.
\end{abstract}

\begin{figure}[t]
  \setlength{\abovecaptionskip}{0.1cm}
  \setlength{\belowcaptionskip}{-0.1cm}
  \centering
  \includegraphics[width=8cm]{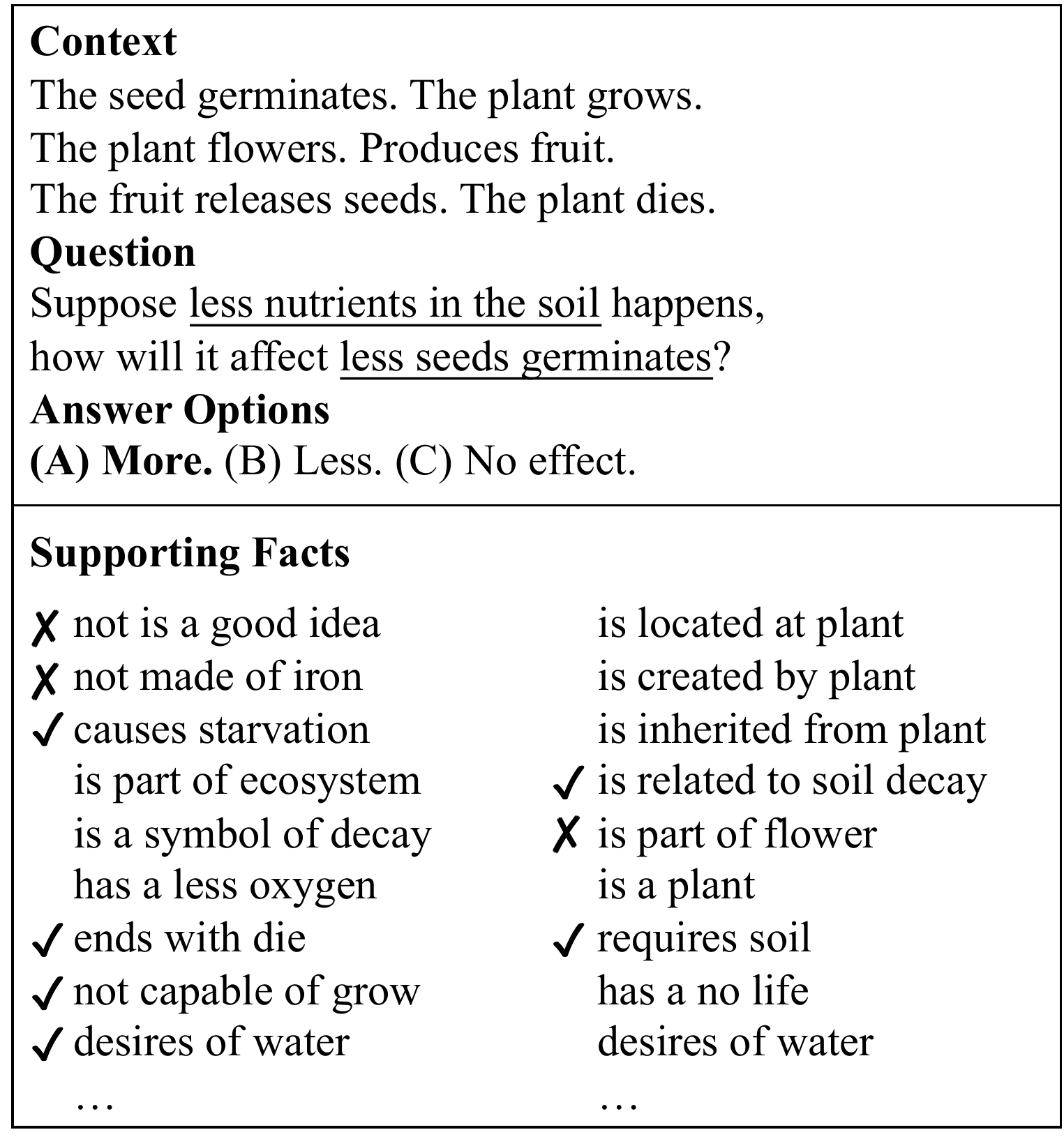}
  \caption{
    \label{fig:intuition}
    (a) An example about supporting facts for a question. The data is from WIQA \cite{tandon2019wiqa} dev set. The supporting facts are generated by COMET \cite{bosselut2019comet}. The quality of the facts is not guaranteed. The facts are mostly semantically related to the \underline{key phrases} in the question, but they contribute differently to answering this commonsense question. For example, ``\textit{is part of flower}'' conveys an attribute of the concept ``\textit{seeds}'', but does not tell us how in fact it will affect ``\textit{less seeds germinates}''. By contrast, ``\textit{causes starvation}'' gives straightforward information that fills the causal gap between ``\textit{less nutrients in the soil}'' and ``\textit{less seeds germinates}''. Therefore, facts like ``\textit{seeds is part of flower}'' do not explain ``\textit{the cause of seeds germination}'' or ``\textit{the effect of nutrients in the soil to the seeds germination}'' that answers the question, whereas ``\textit{causes starvation}'' as an evidence is favorable. (b) The facts with X marks are erased by our proposed REM-Net model, whereas those with check marks survive the multi-hop refinement.
    }
\end{figure}

\section{Introduction}
Commonsense question answering (commonsense QA) is recently an attractive field in that it requires systems to understand the common sense information beyond words, which are normal to human beings but nontrivial for machines. There are plenty of datasets that are proposed for this purpose, for instance, CommonsenseQA \cite{talmor2019commonsenseqa}, CosmosQA \cite{huang2019cosmos}, WIQA \cite{tandon2019wiqa}. Different from traditional machine reading comprehension (MRC) tasks such as SQuAD \cite{rajpurkar2016squad} or NewsQA \cite{trischler2016newsqa} that the key information for answering the questions is directly given by the context paragraph, solving commonsense questions requires a more comprehensive understanding of both the context and the relevant common knowledge, and further reasoning out the hidden logic between them. There are varieties of knowledge bases that meet the need, including text corpora like Wikipedia, and large-scale knowledge graphs \cite{speer2017conceptnet, NELL-aaai15, sap2019atomic}.

Recent popular solution resorts to external supporting facts from such knowledge bases as evidence, to enhance the question with commonsense knowledge or the logic of reasoning \cite{devlin2019bert, liu2019roberta, lv2019graph, lin2019kagnet,xu2020DRLtext}. However, the quality of the supporting facts is not guaranteed, as some of them are weak in interpretability so that do not help the question answering. Specifically, current methods are mainly two-fold. The first group of methods \cite{devlin2019bert, liu2019roberta, bosselut2019comet} pre-train language models on those external supporting facts (e.g., Wikipedia, ConceptNet) so that the models could remember some of the common knowledge, which is empirically proven by Tandon et al. \shortcite{tandon2019wiqa} and Trinh and Le \shortcite{trinh2018do}. The second group of methods \cite{lv2019graph, lin2019kagnet,cao2019bag} incorporates the question with knowledge subgraphs or paths that carry information such as relation among concepts or show multi-hop reasoning process. The structured information is typically encoded via graph models such as GCN \cite{kipf2016semi}, and after which merged with the question features. Generally, current methods all handle evidence by brute force, without further selection or refinement according to the interpretability of the supporting facts. But as the example shown in Figure~\ref{fig:intuition}, some of the supporting facts do not interpret the question, regardless that they are semantically related. Thus, there is need for models that will further our processing of the evidence.

In this paper, we introduce a new recursive erasure memory network (REM-Net) that further refines the candidate supporting fact set. The REM-Net consists of three main components: a query encoder, an evidence generator, and a novel recursive erasure memory (REM) module. Specifically, the query encoder is a pre-trained encoder that encodes the question. The evidence generator is a pre-trained generative model that produces candidate supporting facts based on the question. Compared with those retrieved supporting facts, the generated facts provides new question-specific information beyond the existing knowledge bases. The REM module refines the candidate supporting fact set by recursively matching the supporting facts and the question in feature space to estimate each fact's quality. This estimation helps both updating the question feature and the supporting fact set. The question feature is updated by a residual term, whereas the supporting fact set is updated by removing the low-quality facts. Compared with the standard attention mechanisms \cite{xu2015show,vaswani2017attention} that allocate weights to the supporting facts once, the multi-hop operation in REM module widens the gap of how much each supporting fact contributes to the question answering by the number of recursive steps their features are incorporated for the feature update. Therefore this procedure leads to a refined use of given supporting facts.

We conduct experiments on two commonsense QA benchmarks, WIQA \cite{tandon2019wiqa} and CosmosQA \cite{huang2019cosmos}. The experimental results demonstrate that REM-Net outperforms current methods, and the refined supporting facts are more qualified for the questions. Our contributions are mainly three-fold:

\begin{itemize}
\item We propose a model named recursive erasure memory network (REM-Net) towards evidence refinement according to the commonsense question, which improves the explainability of the supporting facts.
\item We design a new REM module that recursively erases the unqualified supporting facts to provide refined appropriate evidence.
\item Our experimental results demonstrate the superiority of REM-Net compared with other methods that uses external evidence. Moreover, case study shows the interpretability of the refined evidence. 
\end{itemize}

\section{Related Works}
\paragraph{Commonsense Question Answering} 
Similar to open-domain question answering tasks \cite{rajpurkar2018know, kwiatkowski2019natural}, commonsense question answering \cite{tandon2019wiqa, huang2019cosmos} requires open-domain information to support the answer prediction. But different from open-domain question answering tasks that the text comprehension is straightforward and the retrieved open-domain information is direct to the questions, in commonsense question answering tasks the open-domain information is more complicated in that they play a role as evidence to bridge the understanding gap in the commonsense questions. Current works leverage the open-domain information by whether incorporating external knowledge as evidence or training the models to generate evidence. \citet{lv2019graph} extracts knowledge from ConceptNet \citep{speer2017conceptnet} and Wikipedia, and learns features with GCN \citep{kipf2016semi} and graph attention \cite{velickovic2017graph}.   \citet{zhong2019improving} retrieves ConceptNet  \cite{speer2017conceptnet} triplets and train two functions to measure direct and indirect connections between concepts. \citet{rajani2019explain} train a GPT \cite{zhong2019improving} to generate reasonable evidence for the questions. During evaluation, the model generates evidence and predicts the multi-choice answers concurrently. \citet{ye2019align} automatically constructs a commonsense multi-choice dataset from ConceptNet triplets. However, the retrieved or generated evidence are usually not further refined, and some of them could be unnecessary or even confounding to answering the questions. The proposed model explores to refine the original evidence to discover those most supporting evidence to the commonsense questions and therefore provides stronger interpretations.

\paragraph{Memory Networks}
Memory networks \cite{weston2015memory,bordes2015large,miller2016key,sukhbaatar2015end} are proposed to solve early reasoning problems such as bAbI \cite{weston2016towards}) that requires to locate useful information for answer prediction. The sentences are stored into memory slots and later selected for the question answering. Recently, multi-head attention memory networks \cite{dai2019multi} are proposed so that takes advantage of the transformer-based networks. Our proposed model is based on multi-head attention memory network that is modified with a recursive erasure manipulation to adapt to the commonsense question answering tasks for accurate evidence refinement.

\begin{figure}[t!]
  \centering
  \includegraphics[width=.49\textwidth]{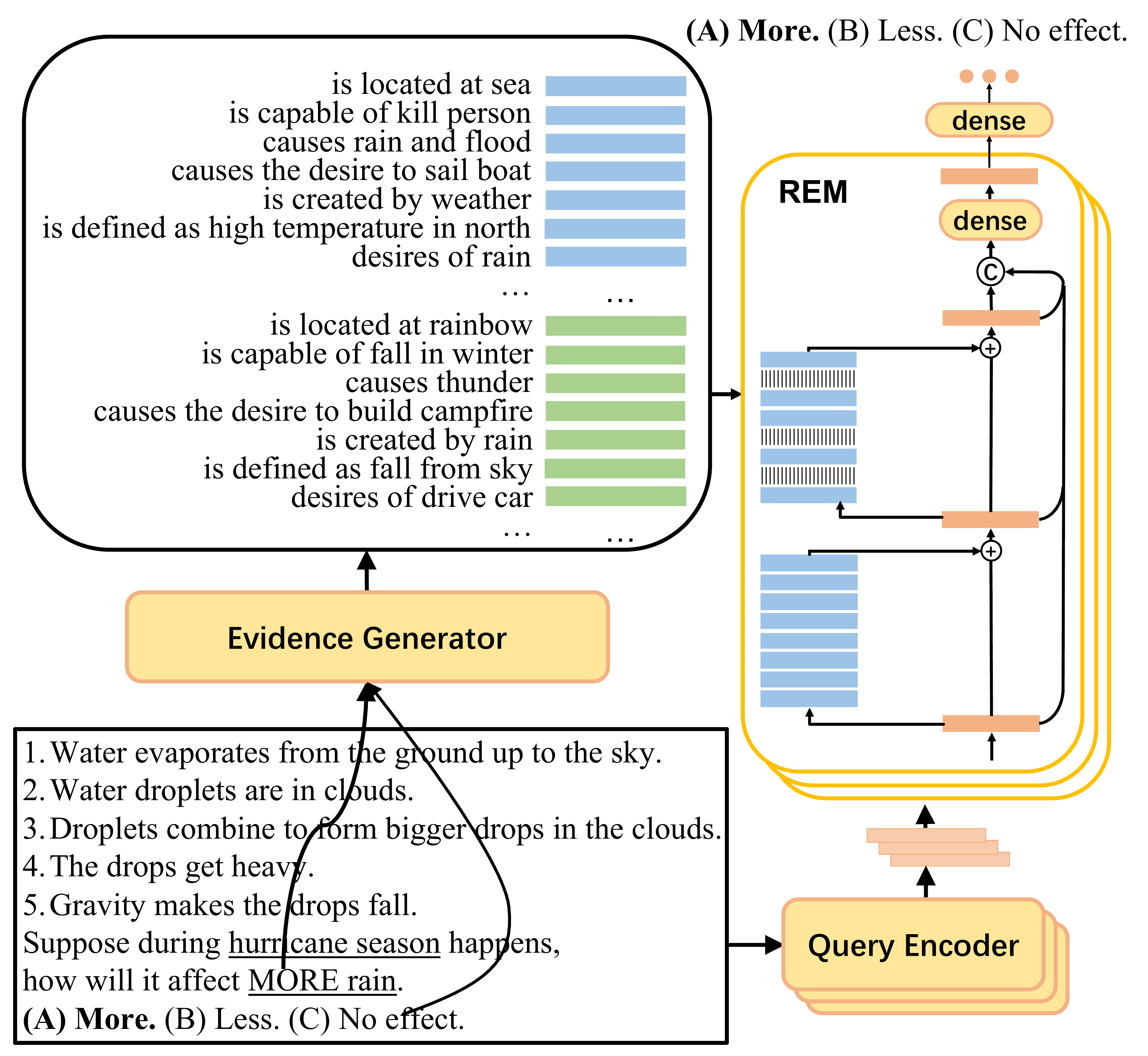}
  \caption{ \label{fig:remnet}  
  The proposed REM-Net with three main components:  
  a query encoder that encodes the commonsense question; 
  an evidence generator providing candidate evidence set in a generation manner; 
  a recursive erasure memory (REM) module that conducts the evidence refinement.
  }
\end{figure}

\section{Recursive Erasure Memory Network}
The main purpose of this model is to refine supporting facts so that they are more explainable to the question. The idea is to recursively erase the unqualified supporting facts. As a result, during the recursive procedure, the retained supporting facts are repeatedly used for updating the features.

The architecture of our model is shown in Figure~\ref{fig:remnet}. It has three main modules. A query encoder encodes the question to a query embedding. An evidence generator produces candidate supporting fact set, and encodes them into embeddings. A recursive erasure memory (REM) module refines the parameterized supporting facts by filtering out unqualified items conditioning on the query embedding. 

\subsection{Query Encoder}
\label{sec:QP}
We follow baselines to use pre-trained language models (e.g., BERT \cite{devlin2019bert}, RoBERTa \cite{liu2019roberta}) to encode the question to contextual embeddings. Given a question as a triplet of (context paragraph, question sentence, answer options), the input sequence is in such format ``\texttt{[CLS] context [SEP] question [SEP] answer option}'', where ``\texttt{[CLS]}'' and ``\texttt{[SEP]}'' are special tokens for pre-trained language model. The output \texttt{[CLS]} embeddings are provided as query to the recursive erasure memory (REM) module.

\begin{figure}[t]
  \centering
  \includegraphics[width=.48\textwidth]{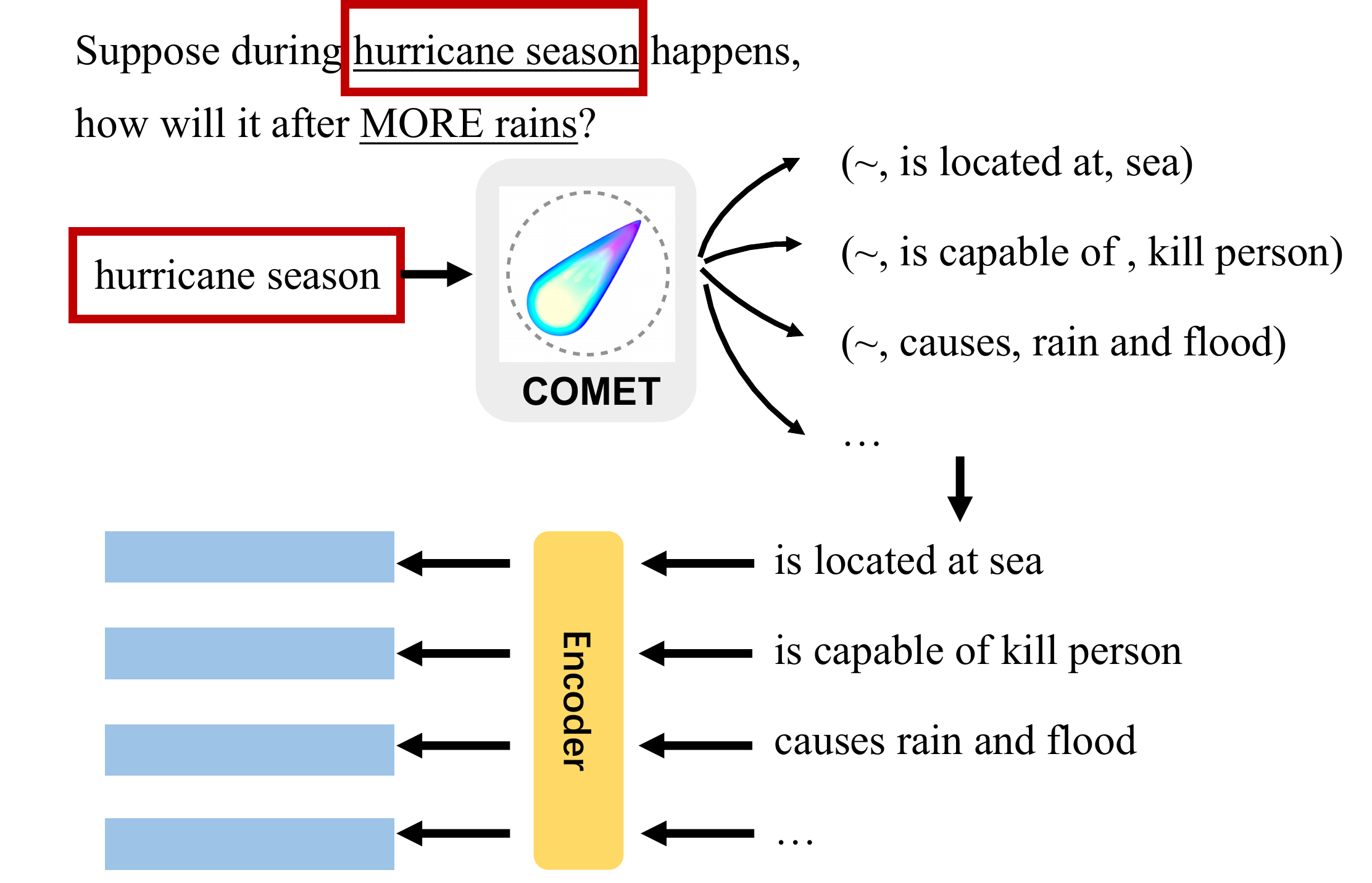}
  \caption{\label{fig:evid_generator}
  Details in the evidence generator. 
  Key phrases are first extracted from the question with rules, then taken as triplet heads to generate relations and triplet tails by COMET \cite{bosselut2019comet}. The triplets are turned into sentences, and finally encoded into evidence embeddings with a pre-trained encoder.
  }
\end{figure}

\subsection{Evidence Generator}
\label{sec:EG}
Generally, for a commonsense question, its supporting facts can be obtained in three main sources: (1) retrieved texts/triplets from knowledge bases, (2) texts/triplets that are generated conditioning on the question, (3) reuse of the context paragraph. Among the three approaches, retrieval-based methods are widely used \cite{lv2019graph, lin2019kagnet}, whereas generation-based methods are barely explored. However, generated supporting facts provide new information that is beyond the commonsense question and knowledge bases. Therefore in this work we use generated supporting facts. We also compare the three sources of supporting facts in the experiment section.

The mechanism of the evidence generator are presented in Figure~\ref{fig:evid_generator}. The generation is achieved by four steps. First, it extracts key phrases from the question.
Second, taking the key phrases as head concepts, it generates relations and tail concepts to complete ConceptNet-like triplets. This is implemented with COMET \cite{bosselut2019comet}, a pre-trained model that is capable of generating commonsense knowledge triplets. Since the generation is based on the key phrases extracted from the question, the generated knowledge triplets are closely related to the question, but the combination of relations and concepts can be new to the existing knowledge bases. Third, the triplets are then converted into natural sentences according to COMET templates\footnote{https://mosaickg.apps.allenai.org/}. 
Finally, the sentences are encoded into embeddings with a pre-trained encoder.

\begin{figure}[t]
  \setlength{\abovecaptionskip}{0.1cm}
  \setlength{\belowcaptionskip}{-0.1cm}
  \centering
  \includegraphics[width=.47\textwidth]{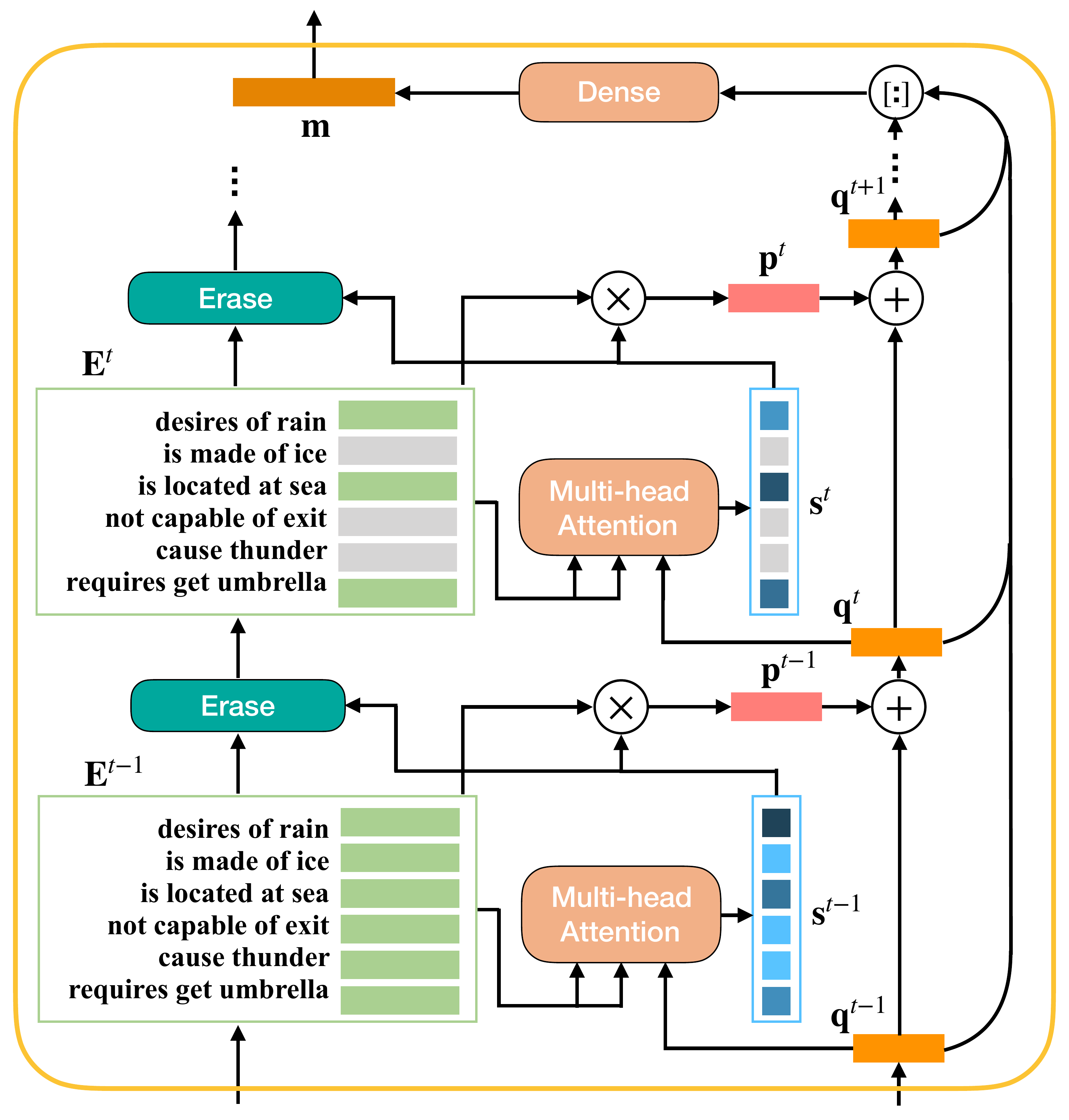}
  \caption{\label{fig:rem}  
  The recursive erasure memory (REM) module. (a) At each recursive step: (i) Multi-head attention \cite{vaswani2017attention} is applied to the evidence score estimation conditioning on the query embedding; (ii) A residual term is calculated by a cross product between the evidence matrix and the query embedding, after which is used for updating the query embedding; (iii) An erasure manipulation is conducted on the evidence set, where the supporting facts with low evidence score are filter from the evidence set. (b) At the end of the overall recursive procedure, the query embeddings at each recursive step are merged by concatenation and a linear projection.}
\end{figure}

\subsection{Recursive Erasure Memory Module}
\label{sec:REM}
The recursive erasure memory (REM) module takes the query embedding and the evidence matrix as input, producing an output feature that merges the updated embeddings. The detailed mechanism is shown in Figure~\ref{fig:rem}. Similar to end-to-end memory networks \cite{sukhbaatar2015end}, REM module matches the question embedding and the evidence matrix recursively to find significant information for the question. However, the manipulations are essentially different. Instead of taking inner product twice on separate evidence matrix to update the context information, REM module uses a single evidence matrix as both input and output, and conducts multi-head attention for information matching. The attention weights are then taken for updating both the evidence matrix and the query embedding.

Given the initial query embedding $\mathbf{q}^0$ from the query encoder and the initial evidence matrix $\mathbf{E}^0$ from the evidence generator, the first recursive step starts with a multi-head attention \cite{vaswani2017attention} that matches both information, so that each supporting fact is allocated with a weight as its evidence score:
\begin{equation}
\label{eq:score_init}
\mathbf{s}^0 = \text{MultiHead}(\mathbf{q}^0, \mathbf{E}^0, \mathbf{E}^0).
\end{equation}
The scores $\mathbf{s}^0$ are then used for updating both the query embedding $\textbf{q}^0$ and the evidence matrix $\mathbf{E}^0$. The query embedding is updated with a residual term that is the outer product of the evidence score vector and the evidence matrix. Meanwhile, the evidence matrix is updated by erasing the low-ranked supporting facts that are sorted by the scores.
The updated query embedding and the updated evidence matrix are then fed into the next recursive step. This procedure is recursively conducted until termination.

We formalize the manipulation at recursive step $t-1$. Current updated query embedding  $\mathbf{q}^{t-1} \in \mathbb{R}^{h}$ and updated evidence matrix $\mathbf{E}^{t-1} \in \mathbb{R}^{I\times h}$ are fed into multi-head attention,
where $h$ is the embedding size and $I$ is the number of stored supporting facts.
$\mathbf{E}^{t-1}$ performs as the key and value and $\mathbf{q}^{t-1}$ as the query.
We obtain evidence scores $\mathbf{s}^{t-1} \in \mathbb{R}^{I}$ for each supporting fact:
\begin{equation} 
\label{eq:score}
\mathbf{s}^{t-1} = \text{MultiHead}(\mathbf{q}^{t-1}, \mathbf{E}^{t-1}, \mathbf{E}^{t-1}).
\end{equation}
The query embedding is updated with a residual term $\mathbf{p}^{t-1}$. It is the outer product of the evidence matrix $\mathbf{E}^{t-1}$ and the evidence score $\mathbf{s}^{t-1}$:
\begin{equation}
\label{qa:newquery}
\begin{split}
\mathbf{p}^{t-1} = {\mathbf{E}^{t-1}}^{\top}\mathbf{s}^{t-1}, \\
\mathbf{q}^{t} = \mathbf{q}^{t-1} + \mathbf{p}^{t-1}.
\end{split}
\end{equation}
The evidence matrix $\mathbf{E}^{t-1}$ is then updated with an erasure manipulation. According to the evidence scores, the supporting facts are sorted, and embeddings of the lowest $k$ supporting facts are removed from the matrix. The evidence matrix is then updated to $\mathbf{E}^t$:
\begin{equation}
\label{eq:newevid}
\mathbf{E}^t = {\left[ \begin{array}{c}
 \mathbf{e}_0^t \\ 
 \mathbf{e}_1^t \\
 \vdots \\ 
 \mathbf{e}_I^t \\
 \end{array} \right]}, 
 \text{   } \mathbf{e}_i^t = 
 \begin{cases}
 \mathbf{e}_i^{t-1},\text{  }s_i^{t-1} \ge s_{[I-k]}^{t-1},\\ \\
 \mathbf{0},\text{  }s_i^{t-1} < s_{[I-k]}^{t-1},
 \end{cases}
\end{equation}
\noindent where $s_{[I-k]}^{t-1}$ is the lowest $k$th score among $\mathbf{s}^{t-1}$.

The resulting query $\mathbf{q}^t$ and evidence $\mathbf{E}^t$ are the inputs of the next recursive step. Therefore, the survived supporting facts are continually matched with the question, whereas the erased supporting facts stop contributing to this procedure. As a consequence, this multi-hop erasure manipulation provides more accurate and interpretable reasoning to the question answering, as the supporting facts are gradually refined.

At the end of the recursive procedure, queries in all recursive steps $\mathbf{q}^t, t\in\{0, 1, ..., T\}$ are concatenated and fed into a fully connected layer,
as the output of the REM module:
\begin{equation}
\label{eq:merge}
\mathbf{m} = [\mathbf{q}^{0}; ...; \mathbf{q}^{T}]\mathbf{W}_m + \mathbf{b}_m,
\end{equation}
\noindent where $[;]$ indicates the concatenation operation, $\mathbf{m} \in\mathbb{R}^{h}$, $\mathbf{W}_m\in\mathbb{R}^{hT\times h}$, and $\mathbf{b}_m \in \mathbb{R}^{h}$.

\subsection{Answer Prediction} 
\label{sec:AP}
The probabilities $Pr$ of choosing the final answer option are: 
\begin{equation}
Pr = \text{SoftMax}([\mathbf{m}_1; ...; \mathbf{m}_C]\mathbf{W}_p + b_p),
\end{equation}
\noindent where $[;]$ indicates concatenation, 
\{$\mathbf{m}_1$, ..., $\mathbf{m}_C$\} are outputs of the REM module for each answer option,
and $C$ is the number of answer options. 
$\mathbf{W}_p \in \mathbb{R}^{h\times 1}$, $b_p \in \mathbb{R}$.

\section{Experiments}
We evaluate REM-Net on two commonsense QA datasets, WIQA \cite{tandon2019wiqa} and CosmosQA \cite{huang2019cosmos}. We then conduct ablation study on the REM module, and show several cases of REM-Net's evidence refinement.

\subsection{Data}
\paragraph{WIQA} \cite{tandon2019wiqa} contains counterfactual questions in such a fixed pattern as ``\textit{suppose ... happens, how will it affects ...}'', in which the two clauses relate to cause and effect separately. The context paragraphs provide descriptions of natural phenomenons, which are manually written based on specifically defined ``influence graphs''. The questions are split into three types (``in-para'', ``out-of-para'', ``no-effect'') depending on whether the questions are derived from the original ``influence graphs''. For ``out-of-para'' and ``no-effect'' questions, the context paragraphs are irrelevant to the questions, so that they are unable to provide meaningful evidence. \paragraph{CosmosQA} \cite{huang2019cosmos} includes questions of daily life scenarios, such as cultural norms, counterfactual reasoning, situational fact, and temporal event. The scenarios are plentiful and the questions are also diverse. The questions are in a multi-choice format.

\begin{table}[t!]
  \setlength{\belowcaptionskip}{-0.2cm}
  \small
  \centering
  \begin{tabular}{
  p{.225\textwidth}
  p{.027\textwidth}<{\centering}
  p{.027\textwidth}<{\centering}
  p{.027\textwidth}<{\centering}
  p{.03\textwidth}<{\centering}
  }
  \toprule
  \textbf{Method} & \textbf{In} & \textbf{Out} & \textbf{No} & \textbf{Total} \\
  \midrule
  \textit{Baselines} \\
  \quad Majority \shortcite{tandon2019wiqa}$^{*}$ & 45.46 & 49.47 & \,\,\,0.55 & 30.66 \\
  \quad Polarity \shortcite{tandon2019wiqa}$^{*}$ & 76.31 & 53.59 & \,\,\,0.27 & 39.43 \\
  \quad Adaboost \shortcite{freund1995desicion}$^{*}$ & 49.41 & 36.61 & 48.42 & 43.93 \\
  \quad Decomp-Attn \shortcite{parikh2016a}$^{*}$ & 56.31 & 48.56 & 73.42 & 59.48 \\
  \midrule
  \textit{Implicit use of evidence} \\
  \quad \bertbase (no para) & 66.60 & 64.29 & 74.90 & 69.13 \\
  \quad \bertbase & 70.57 & 58.54 & 91.08 & 74.26 \\
  \quad \bertbase (ensemble) & 71.51 & 61.82 & 90.72 & 75.61 \\
  \quad \bertlarge & 73.40 & 63.88 & 90.52 & 76.69 \\
  \quad \bertlarge (ensemble) & 71.51 & 62.73 & 90.04 & 75.69 \\
  \quad \robertalarge & 78.87 & 73.48 & 88.69 & 80.79 \\
  \quad \robertalarge (ensemble) & 77.46 & 71.39 & 90.48 & 80.44 \\
  \midrule
  \textit{Explicit use of evidence} \\
  \quad MemN2N \shortcite{sukhbaatar2015end} & 38.50 & 38.01 & 39.52 & 38.85 \\
  \quad Input Aug (\bertbase) & 70.57 & 61.00 & 90.72 & 75.12 \\
  \quad Input Aug (\bertlarge) & 73.40 & 63.88 & 90.52 & 76.69 \\
  \quad Input Aug (\robertalarge) & 75.66 & 71.59 & 90.60 & 80.25 \\
  \quad SDP Att (\bertbase) \shortcite{vaswani2017attention} & 72.83 & 63.71 & 63.71 & 75.26 \\
  \quad SDP Att (\bertlarge) & 72.26 & 66.26 & 90.28 & 77.36 \\
  \midrule
  \textit{Ours} \\
  \quad REM-Net (\bertbase) & 70.94 & 63.22 & 91.24 & 76.29 \\
  \quad REM-Net (\bertlarge) & 73.21 & 68.14 & 90.84 & 78.52 \\
  \quad REM-Net (\robertalarge) & 76.23 & 69.13 & 92.35 & 80.09 \\
  \bottomrule
  \end{tabular}
  \caption{\label{tab:wiqa} 
  Results (accuracy\%) on the WIQA test set, including accuracies on three separate question types (In=``in-para'', Out=``out-of-para'', No=``no-effect''), and the overall test set. The baselines labeled with $^{*}$ are taken from \citet{tandon2019wiqa}, in which the used test set is slightly different.
 }
\end{table}

\begin{table}[t!]
  \small
  \centering
  \begin{tabular}{
  p{.32\textwidth}
  p{.07\textwidth}<{\centering}
  }
  \toprule
  \textbf{Method} & \textbf{Dev} \\
  \midrule
  \textit{Baselines} \\
  \quad Sliding Window \shortcite{richardson2013mctest} & 25.0 \\
  \quad Stanford Attentive Reader \shortcite{chen2016a} & 45.3 \\
  \quad Gated-Attention Reader \shortcite{dhingra2017gated} & 46.9 \\
  \quad Co-Matching \shortcite{wang2018a} & 45.9 \\  
  \midrule
  \textit{Implicit use of evidence} \\
  \quad Commonsense-Rc \shortcite{wang2018yuanfudao} & 47.6 \\
  \quad GPT-FT \shortcite{radford2018improving} & 54.0 \\
  \quad DMCN \shortcite{zhang2019dcmn} & 67.1 \\
  \quad \bertlarge \shortcite{devlin2019bert} & 66.2 \\
  \quad \bertlarge (ensemble) & 67.1 \\
  \quad \bertlarge (multiway) \shortcite{huang2019cosmos} & 68.3 \\
  \quad \robertalarge \shortcite{liu2019roberta} & 78.6 \\
  \midrule
  \textit{Explicit use of evidence} \\
  \quad MemN2N \shortcite{sukhbaatar2015end} & 30.6 \\
  \quad Input Aug (\bertlarge) & 67.1 \\
  \quad Input Aug (\robertalarge) & 80.8 \\
  \quad SDP Att (\bertlarge) & 27.4 \\
  \quad SDP Att (\robertalarge) & 25.6 \\
  \midrule
  \textit{Ours} \\
  \quad REM-Net (\bertlarge) & 69.5 \\
  \quad REM-Net (\robertalarge) & 81.4 \\
  \bottomrule
  \end{tabular}
  \caption{\label{tab:cosmosqa}
  Results (accuracy\%) on the CosmosQA dev set. 
  }
\end{table}

\subsection{Compared Methods}
\label{sec:baselines}
We compare the performance of REM-Net with several groups of competitive methods.

\paragraph{Group 1:} 
Baselines. For WIQA, Majority predicts the most frequent answer option in the training set. Polarity predicts answers with the most comparative words. Adaboost \cite{freund1995desicion} uses bag-of-words features in the questions. Decomp-Attn \cite{parikh2016a} is a decomposable attention model that computes attention between sentences. For CosmosQA, Sliding Window \cite{richardson2013mctest} evaluates the similarity between context paragraph and answer options. Stanford Attentive Reader \cite{chen2016a}, Gated-Attention Reader \cite{dhingra2017gated} and Co-Matching \cite{wang2018a} are reading comprehension systems that performs attention mechanism differently. 

\paragraph{Group 2:} 
Implicit incorporation of supporting evidence. Commonsense-RC \citep{wang2018yuanfudao} is an LSTM-based model pre-trained on RACE \cite{lai2017race}. Transformer-based pre-trained language models such as BERT \cite{devlin2019bert} and RoBERTa \cite{liu2019roberta} that learn from large scale corpora. 

\paragraph{Group 3: } 
Explicit use of supporting evidence.
End-to-end memory networks (MemN2N) \cite{sukhbaatar2015end} are LSTM-based recursive models that recursively match the context to the question. Input augmentation (Input Aug) directly augments the question by appending the supporting evidence to the question text and encodes them to contextual embeddings with pre-trained language models. Scaled dot-product attention (SDP Att) \cite{vaswani2017attention} allocates attention weights to each supporting facts.
In our experiments, the evidence is the same as REM-Net, which are supporting facts generated by the evidence generator.

\subsection{Experimental Setup}
\paragraph{Seed Key Phrases Extraction}
The supporting facts are generated based on the key phrases. For WIQA, we set a rule to extract those key phrases. Since each of the question sentences consists of a cause clause and an effect clause with fixed pattern, we remove the pattern words to obtain two groups of key phrases, and separately generate two groups of supporting facts. For CosmosQA, we use the TAGME \cite{assante2019enacting} toolkit~\footnote{https://tagme.d4science.org/tagme/} to automatically tag the key phrases from the context paragraphs and the question.

\paragraph{Implementation Details}
We use BERT \cite{devlin2019bert} and RoBERTa \cite{liu2019roberta} as the backbones. The sequence length for the query encoder is 128, which is sufficient to include the input sequence ``\texttt{[CLS] context [SEP] question [SEP] answer option}'' ($>88\%$). For the evidence generator, the sequence length is set to 30 and covers the vast majority of evidence sentences ($>99\%$).

For experiments on WIQA, since there are two groups of supporting facts relating to the cause and the effect, we adopt two parallel REM modules to separately refine them. The model is optimized by Adam \cite{kingma2015adam} with a learning rate of $1\times 10^{-5}$. Warmup steps are 1000. We train 25 epochs with batch size 8. For the termination condition of the recursion, we set a fixed recursive step to 2. The upper bound of erased evidence sentences at each recursive step is 50. For CosmosQA, we use a single REM module to refine the evidence. The model is optimized using the Adam optimizer with a learning rate of $5\times 10^{-6}$ and warmup steps of 1500. The model is trained with 10 epochs and a batch size of 4. The fixed recursive step is 2. The upper bound of erased evidence sentences at each recursive step is 10. 

\begin{table}[t!]
  \setlength{\abovecaptionskip}{0.1cm}
  \setlength{\belowcaptionskip}{-0.1cm}
  \small
  \centering
  \begin{tabular}{
  p{.18\textwidth}
  p{.04\textwidth}<{\centering}
  p{.04\textwidth}<{\centering}
  p{.04\textwidth}<{\centering}
  p{.04\textwidth}<{\centering}
  }
  \toprule
   & \textbf{In} & \textbf{Out} & \textbf{No} & \textbf{Total} \\
  \midrule
  REM-Net (\bertlarge) & 73.21 & 68.14 & 90.84 & 78.52 \\
  \quad\quad w/o E & 69.81 & 52.79 & 91.79 & 72.09 \\
  \quad\quad w/o E, w/o R & 62.08 & 43.27 & 92.35 & 67.10 \\
  \bottomrule
  \end{tabular}
  \caption{\label{tab:ablation_wiqa} 
  Ablation study on REM-Net (\bertlarge) that are conducted on WIQA. E signifies the erasure manipulation, whereas R indicates to the recursive mechanism. In=``In-para'' type, Out=``Out-of-para'' type, No=``No-effect'' type.
 }
\end{table}

\begin{table}[t!]
  \setlength{\abovecaptionskip}{0.1cm}
  \setlength{\belowcaptionskip}{-0.1cm}
  \small
  \centering
  \begin{tabular}{
  p{.23\textwidth}
  p{.08\textwidth}<{\centering}
  p{.08\textwidth}<{\centering}
  }
  \toprule
   & \textbf{Dev} & \textbf{Test} \\
  \midrule
  REM-Net (\bertlarge) & 69.49 & 70.07 \\
  \quad\quad\quad w/o E & 68.44 & 68.58 \\
  \quad\quad\quad w/o E, w/o R & 68.27 & 68.53 \\
  \bottomrule
  \end{tabular}
  \caption{\label{tab:ablation_cosmosqa} Ablation study on REM-Net (\bertlarge) that are conducted on CosmosQA. E denotes the erasure manipulation, while R refers to the recursive mechanism.
  }
\end{table}

\subsection{Experimental Results}
The experimental results are presented in Table~\ref{tab:wiqa} and Table~\ref{tab:cosmosqa}. 
The REM-Net is compared with three groups of methods. 
It is shown that the REM-Net outperforms the compared approaches in most of the experiments. Besides, models perform differently on different data. In the CosmosQA dataset, our REM-Net outperforms all of the compared methods. In WIQA, REM-Net (\bertlarge) is superior, whereas REM-Net (\robertalarge) is comparable to other methods. REM-Net (\robertalarge) is mainly inferior in the ``in-para'' and ``out-of-para'' data type, but surpasses compared methods in the ``no-effect'' data type. This is because the majority of the “in-para” and “out-of-para” evidence is meaningful to the question, and thus the erasure operation from the REM module provides limited effect.

\subsection{Ablation Study} 
We further investigate the details in REM-Net. The results are shown in Table~\ref{tab:ablation_wiqa} and Table~\ref{tab:ablation_cosmosqa}. It is observed that removing the erasure manipulation from the REM module leads to performance drop. 
This indicates that excluding those low-quality supporting facts benefits the results. Further removing the recursive mechanism, which means the REM module calculates the evidence scores once, brings a further performance drop. This indicates that recursively estimating the evidence sentences refines the understanding of the question and provides better interpretation. Therefore, erasure manipulation and the recursive mechanism both contribute to the benefits provided by our model.

\begin{figure}[t]
    \setlength{\abovecaptionskip}{0.1cm}
    \setlength{\belowcaptionskip}{-0.1cm}
    \centering
    \includegraphics[width=8.5cm]{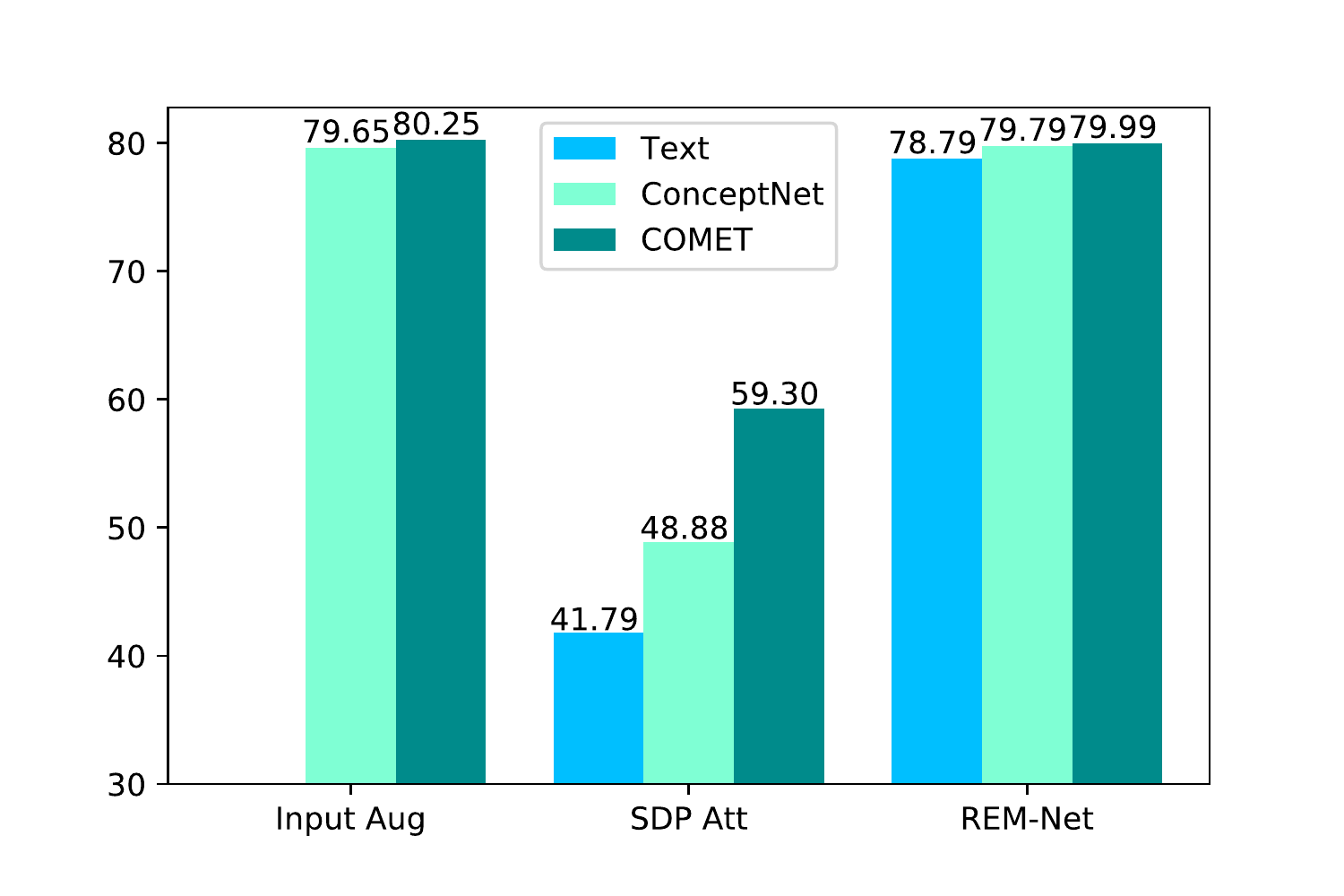}
    \caption{\label{fig:compare_source} Comparison accuracies (\%) on WIQA test set among three evidence sources. The base model being used for the three methods is \robertalarge.}
\end{figure}

\begin{figure*}[t]
    \setlength{\abovecaptionskip}{0.1cm}
    \setlength{\belowcaptionskip}{-0.4cm}
    \centering
    \includegraphics[width=\textwidth]{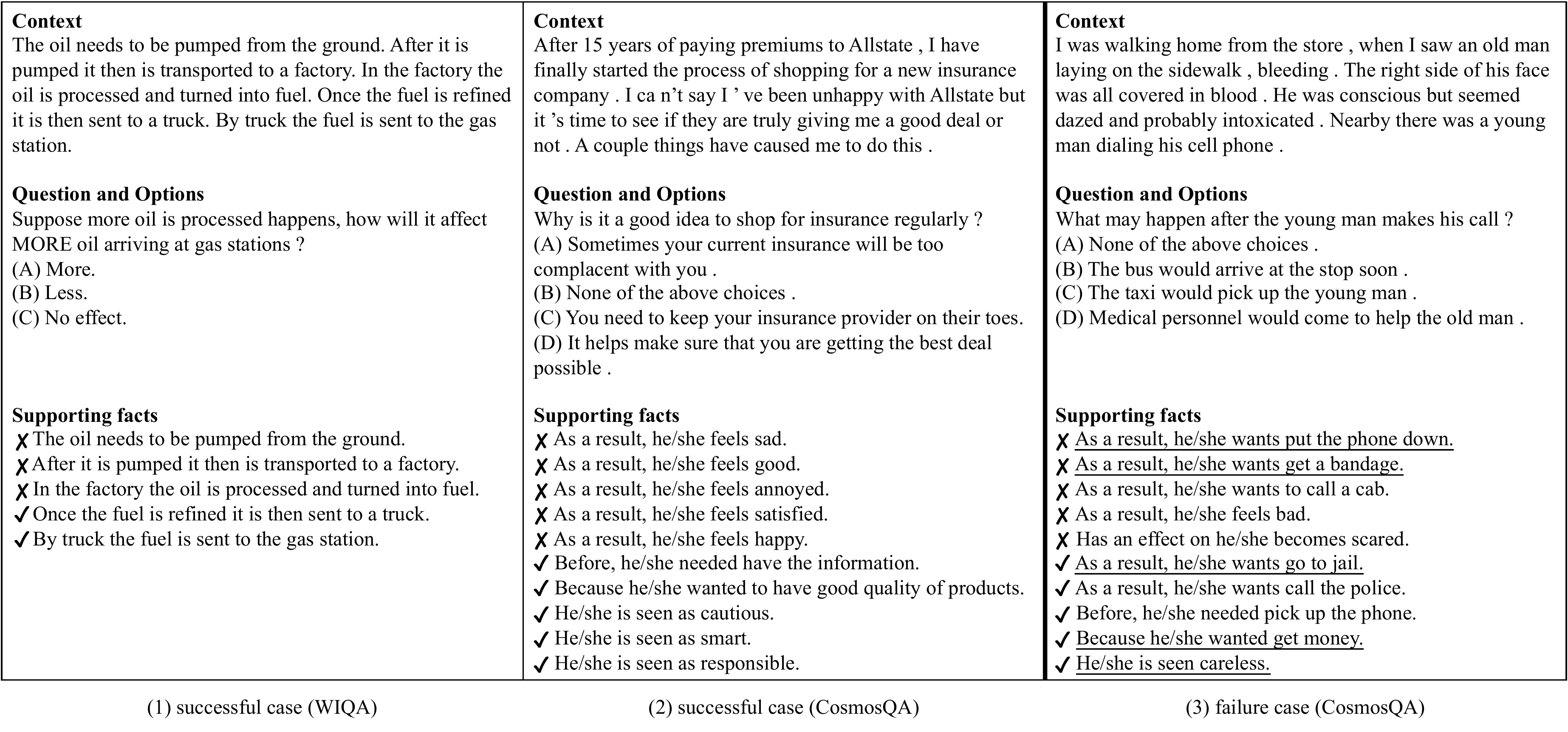}
    \caption{\label{fig:case_study} Examples of evidence refinement by REM-Net. 
    Case (1) presents a successful case from the WIQA test set. The supporting facts are context sentences.
    Case (2) is a successful case from the CosmosQA dev set, in which the presented supporting facts are part of the generated facts by the evidence generator. 
    Case (3) shows a failure case from the CosmosQA dev set. The presented supporting facts are part of the generated facts by the evidence generator, therein the underlined facts are incorrectly erased or retained.
    }
\end{figure*}

\subsection{Generated Evidence Versus Retrieved Evidence}
We compare the quality of generated evidence and retrieved evidence. For a fair comparison, both evidence are based on ConceptNet. Specifically, the generated evidence are produced by COMET that is pre-trained on ConceptNet, whereas the retrieved evidence is directly retrieved from ConceptNet. Besides, to provide baseline results, we simply take the context paragraph provided by the question as another type of evidence. In the experiments, we provide different types of evidence to three methods that use evidence in an explicit manner, which are input augmentation, scaled dot-product attention, and the proposed REM-Net. The comparison results are shown in Figure~\ref{fig:compare_source}. It is shown that in the three methods, incorporating the generated evidence gives better results than the retrieved evidence. The performance gap is especially obvious for scaled dot-product attention, since it selects the evidence once with the attention weights. The REM-Net refines the evidence in a multi-hop manner, and the performance gap between different evidence are small, but generated evidence still gives better result. 

\subsection{Case Study}
We show three cases to see the qualify of refined evidence, as presented in Figure~\ref{fig:case_study}.

Figure~\ref{fig:case_study} (1) shows a successful case in WIQA. The supporting facts are context paragraph sentences. The provided context paragraph covers a whole process of fuel production, whereas the question is about the causal relation between oil processing and fuel transportation.  REM-Net erases the irrelevant oil processing sentences, retaining the sentences about fuel transportation.

Figure~\ref{fig:case_study} (2) presents a successful case in CosmosQA, in which REM-Net refines generated supporting facts. The question is about good reasons for regularly buying insurance. The context paragraph tells a story about the narrator deciding to change his/her insurance products, but the reason for his/her decision is not provided. The generated facts supply such reasons. The erased facts such as ``\textit{As a result, he/she feels sad}'' or ``\textit{As a result, he/she feels happy}'' do not interprets the question, since changing the insurance products are normally someone's rational decision. On the contrary, ``\textit{Because he/she wanted to have good quality of products}'' support the question well. It is intuitive that the retained facts interprets the question better.

Figure~\ref{fig:case_study} (3) shows a failure case in CosmosQA. This question is about the follow-up events after the young man makes a call to help the old man. The erasure by REM-Net seems unreasonable.
The erased supporting facts include ``\textit{As a result, he/she wants put the phone down}'' and ``\textit{As a result, he/she wants get a bandage}'', which are events related to the question. On the other hand, the retained supporting facts contain ``As a result, he/she wants go to fail'' and ``Because he/she wanted get money''. Including the context and the question, these supporting facts are unreasonable inferences. This case indicates that the erasure operation of REM-Net does not cover all the questions. One of the reasons is that the commonsense questions are in varied domains so that some of the domains with fewer samples are not well trained. 

\vspace{-2mm}
\section{Conclusion}
In this paper, we propose a recursive erasure memory network (REM-Net) that refines evidence for commonsense question. It recursively estimates quality of each supporting fact based on the question, and refines the supporting fact set accordingly. The recursive procedure leads to repeated use of high-quality supporting facts, so that the question answering is conducted by useful information. Experimental results demonstrates that REM-Net is effective for the commonsense QA tasks, and the evidence refinement is interpretable. Besides, we evaluate the quality of generated evidence compared to retrieved evidence, learning that using generated evidence gives better performance.

\section{Acknowledgments}
The authors would like to thank Yu Cao, Jinghui Qin, Zheng Ye, Zhicheng Yang for their useful discussions.
This work was supported in part by National Natural Science Foundation of China (NSFC) under Grant No.U19A2073 and No.61976233, Guangdong Province Basic and Applied Basic Research (Regional Joint Fund-Key) Grant No.2019B1515120039,  Shenzhen Basic Research Project (Project No. JCYJ20190807154211365), Zhijiang Lab’s Open Fund (No. 2020AA3AB14) and CSIG Young Fellow Support Fund.

\bibliography{remnet}
\bibliographystyle{aaai21}

\end{document}